\documentclass[10pt,twocolumn,letterpaper]{article}

\usepackage{times}
\usepackage{epsfig}
\usepackage{graphicx}
\usepackage{amsmath}
\usepackage{amssymb}
\usepackage{booktabs} 
\usepackage{subcaption} 
\usepackage{url}
\usepackage{fullpage}

\newtheorem{procedure}{Procedure}

\begin{document}
\title{Understanding Unequal Gender Classification Accuracy \\from Face Images}

\author{Vidya Muthukumar,$^{1,2}$ Tejaswini Pedapati,$^1$ Nalini Ratha,$^1$ Prasanna Sattigeri,$^1$ Chai-Wah Wu,$^1$ \\Brian Kingsbury,$^1$ Abhishek Kumar,$^1$ Samuel Thomas,$^1$ \\Aleksandra Mojsilovi\'c,$^1$ and Kush R.\ Varshney$^1$\\\\$^1$IBM Research\\$^2$University of California, Berkeley}

\date{}

\maketitle
\begin{abstract}
Recent work shows unequal performance of commercial face classification services in the gender classification task across intersectional groups defined by skin type and gender. Accuracy on dark-skinned females is significantly worse than on any other group. In this paper, we conduct several analyses to try to uncover the reason for this gap. The main finding, perhaps surprisingly, is that skin type is not the driver. This conclusion is reached via stability experiments that vary an image's skin type via color-theoretic methods, namely luminance mode-shift and optimal transport. A second suspect, hair length, is also shown not to be the driver via experiments on face images cropped to exclude the hair. Finally, using contrastive post-hoc explanation techniques for neural networks, we bring forth evidence suggesting that differences in lip, eye and cheek structure across ethnicity lead to the differences. Further, lip and eye makeup are seen as strong predictors for a female face, which is a troubling propagation of a gender stereotype. 
\end{abstract}

\section{Introduction}

The problem of unequal accuracy rates across groups has recently been highlighted in gender classification from face images.
A study by NIST shows that automated gender classification algorithms are more accurate for males than females~\cite{ngan2015face}. 
Going further, Buolamwini and Gebru created a dataset of parliament members from three European and three African countries --- the Pilot Parliaments Benchmark (PPB), balanced across two attributes: gender and Fitzpatrick skin type \cite{Fitzpatrick1988}, and evaluated the accuracy of three commercial facial gender classifiers \cite{buolamwini2018gender}. All three achieved much lower accuracy on dark-skinned females (Fitzpatrick skin types IV--VI) than light-skinned females, dark-skinned males, and light-skinned males.  (Note that gender classification is a distinct task from race classification \cite{FuHH2014}.)

The discrepancy is conjectured to be largely due to imbalanced training datasets and test benchmarks.
Commonly used training datasets such as CelebA~\cite{liu2015deep} and IMDb face  \cite{Rothe-IJCV-2016} are made up of celebrities and biased towards light-skinned people. 
Test benchmarks such as Labeled Faces in the Wild~\cite{huang2008labeled} and Adience~\cite{eidinger2014age} are also biased \cite{buolamwini2018gender}, so high overall accuracies achieved on these test datasets obfuscate the inequality issue.
The IJB-A dataset purports to be geographically diverse \cite{klare2015pushing}, but a close examination reveals that only 8 percent of the faces are of African descent, whereas more than 50 percent of the faces are of European descent.
The PPB dataset is the first of its kind to be balanced by gender \emph{and} balanced between African and European descent \cite{buolamwini2018gender}.

These works, however, do not investigate the underlying causes of the unequal misclassification rates in gender classification. In particular, since the partition in \cite{buolamwini2018gender} is \textit{phenotypic} into different skin type categories but the dark-skinned people are predominantly of African descent, it may be that other features, such as hairstyle, facial structure, cosmetics or clothing are the reason for disparity, rather than skin type alone \cite{Buolamwini2017}. A study of unequal gender classification accuracy, conducted using images with different parts of the face masked out, points to the nose region as important, but does little to disentangle the various aspects of identity \cite{OzbudakKCG2010}.  Buolamwini points to several shortcomings of that study and calls for ``further scholarship that attends to the impact of phenotypic 
characteristic on gender classification that extends beyond skin type'' \cite{Buolamwini2017}.

Heeding this call, we rigorously analyze gender classifiers and test the extent to which various features influence the classification outcome by skin type and gender. The contributions of this paper are as follows.
\begin{itemize}
\item Using principles from color theory and the framework of optimal transport, we test \textit{stability} to skin type by varying the skin type of a face keeping all other features fixed, and statistically show that the effect of skin tone on classification outcome is minimal.
Thus, the unequal accuracy observed in \cite{buolamwini2018gender} likely arises not specifically because of the skin type, but other correlated features of identity \cite{akerlof2010identity}.

\item Motivated by a visual observation that most misclassified dark-skinned females have short hair, we test the significance of hair patterns in gender classification.
We find that ignoring hair information retains high overall accuracy as well as differential performance, suggesting that information about the hair is also unimportant.

\item Finally, we use recently proposed ideas on contrastive explanation~\cite{dhurandhar2018explanations} to show that neural networks used for gender classification latch on to various facial features like lips, cheeks, and eyes with cosmetics as sufficient explanations for the female gender --- suggesting that discrepancies in these features are the root of the inequality observed in intersectionally-defined groups.
\end{itemize}

\section{Setup}

\subsection{Pilot Parliaments Benchmark dataset}

\begin{table}
\centering
  \caption{Gender and skin type composition of PPB*/PPB dataset.}
  \label{tab:ppbstar}
  \begin{tabular}{cccc}
   \toprule
    Set & Number & Female & Male \\
    \midrule
    All & $1204/1270$ & $42.1/44.6\%$ & $57.9/55.4\%$ \\
    \midrule
    Dark & $507/589$ & $41.8/45.9\%$ & $58.2/54.1\%$ \\
    \midrule
    Light & $697/681$ & $42.4/43.4\%$ & $57.6/56.6\%$ \\
   \bottomrule
  \end{tabular}
\end{table}

The PPB dataset is the first benchmark dataset that is balanced across gender and Fitzpatrick skin type; the methodology of its collection is detailed in \cite{buolamwini2018gender}. The creators intentionally chose countries with majority populations at opposite ends of the skin type scale to make the lighter/darker dichotomy more distinct. The images are uniform in (high) resolution quality, pose, illumination and expression, reducing the possibility of attributing differences in performance to variations in these quantities, all of which are known to be significant technical challenges \cite{sim2002cmu}. 

We use an approximation of the PPB dataset for the experiments in this paper. This dataset contains images of parliament members from the six countries identified in \cite{buolamwini2018gender} and were manually labeled by us into the categories dark-skinned and light-skinned.\footnote{The images were accessed in January 2018. We do not work with the PPB dataset directly due to its terms and conditions of use.} Our approximation to the PPB dataset, which we call PPB*, is very similar to PPB and satisfies the relevant characteristics for the study we perform.
Table \ref{tab:ppbstar} compares the decomposition of the original PPB dataset and our PPB* approximation according to skin type and gender.

\subsection{Classification models}\label{sec:models}


We employ several classifiers in our experiments. The first classifier is the IBM Watson gender classifier service available in August 2018, which achieves $99\%$ accuracy on several test benchmarks, as well as $99\%$ accuracy on the light male, light female and dark male groups of the PPB* dataset.
We access the gender classifier using the API, which takes in an input image of variable size and returns (in the event that a face is detected) a score $s \in [0,1]$ that the image is of a male person. Values $s \le 0.5$ are classified female and values $s > 0.5$ are classified male. Accuracies on the PPB* dataset are presented in Table~\ref{tab:accuracies}.
The accessibility of scores from the IBM Watson API as well as its high level of performance make this a good classifier for carrying out stability experiments. (Scores are not available from the other two commercial classifiers studied in \cite{buolamwini2018gender}.)

The second and third classifier are needed for studying face-only cropped images (i.e.\ cropped so that no hair is present) because the commercial classifier API does not offer the flexibility to restrict modeling to smaller cropped areas.  As a second classifier, we use IBM Watson's ``deep-face-features" API to extract a $256$-dimensional representation for every face-cropped image. We train a downstream support vector machine (SVM) classifier with radial basis function (RBF) kernel  using $20000$ images from the CelebA dataset for training and $5000$ validation images to choose the RBF kernel parameter. The third classifier is the same as the second in terms of the SVM and its training, but has a more modern feature extractor: a convolutional neural network (CNN) trained on the recently created VGGFace2 dataset \cite{CaoSXPZ2018}.
In particular, we use the ResNet-50 network to extract a $128$-dimensional representation for every face-cropped image.\footnote{\url{https://github.com/ox-vgg/vgg_face2}}

\begin{figure}
\includegraphics[height=1in,width=3in]{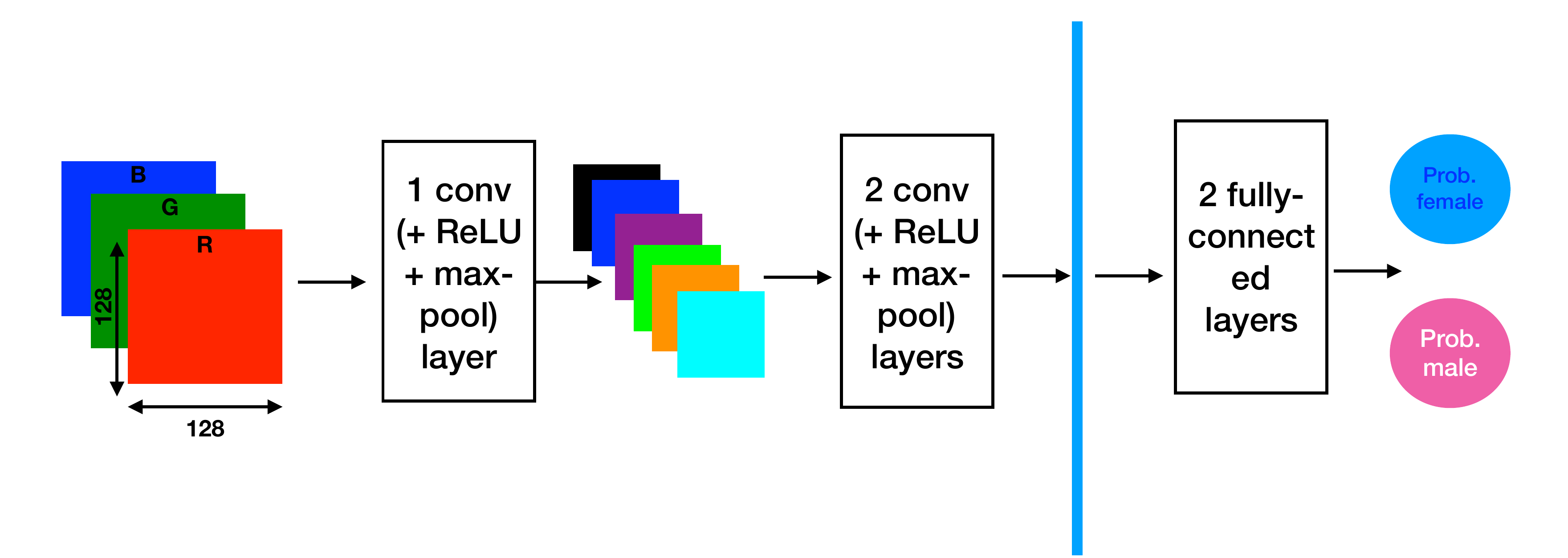}
\caption{Customized classifier trained on CelebA.}\label{fig:cnn}
\end{figure}

The fourth classifier is needed because unfortunately, apart from the scores, we only have black-box access to the Watson API. The details of the model architecture are not available and we cannot inspect any intermediate layers. For interpretability experiments, we create a customized classifier: a simple CNN described in Figure~\ref{fig:cnn}.
We train this neural network on CelebA \cite{liu2015deep} ($162,768$ training images).
Before input to the neural network, all images are face-cropped with padding, eye-aligned and resized to $128 \times 128$. Accuracies on the PPB* dataset for the customized model are presented in Table~\ref{tab:accuracies}.
While they are not as good as the Watson classifier, especially on females, they do achieve close to state-of-the-art accuracy on males, and serve as a model that can be further interrogated.
\begin{table}
\centering
  \caption{Accuracy on PPB* for dark females (DF), light females (LF), dark males (DM), and light males (LM).}
  \label{tab:accuracies}
  \begin{tabular}{ccccc}
    \toprule
    Classifier & DF & DM & LF & LM \\
    \midrule
    Watson & $82.5\%$ & $99.3\%$ & $98.5\%$ & $99.5\%$  \\
    \midrule
    Customized & $70.5\%$ & $95.7\%$ & $86.8\%$ & $97.5\%$ \\
    \bottomrule
  \end{tabular}
\end{table}

\section{Experiments}

The first set of experiments tests the stability of gender classification algorithms to variation in skin type.
Next, we test the influence of hair length, by seeing whether the unequal performance persists when we remove all information about this attribute from faces.
Finally, we seek sufficient explanations on faces for the classification decisions of female and male respectively.

\subsection{Stability experiment: Does skin type \emph{alone} influence gender classification?}

In the first set of experiments, we systematically isolate the skin type and test the gender classification outcome for significant changes as a function of varying skin type.
Isolating a latent facial attribute, and thus changing it, is in general known to be a challenging computer vision task.
Likelihood based generative models~\cite{kingma2018glow} and conditional generative adversarial networks (GANs)~\cite{choi2017stargan,perarnau2016invertible} have made recent progress in varying attributes like hair color and facial expressions. However, these tools themselves are trained on biased celebrity datasets. Moreover, these approaches are not effective in varying one attribute \emph{in isolation}, leaving other attributes unchanged.
We empirically show the existence of an approximately low-dimensional structure in color space that describes the group of human skin types.
Leveraging this structure, we provide simple but mathematically grounded rules to change the skin type of a face.

\subsubsection{A low-dimensional skin type group in $\text{YCrCb}$ space}

Recall that image pixels can be represented in the $3$-dimensional vector space $[0,255]^3$.
Multiple bases for the color space such as the standard RGB~\cite{feitosa2014mathematical}, HSV~\cite{oliveira2009skin} and YCrCb~\cite{hsu2002face}, have been used to create skin detection rules.  More recently, hybrid rules have been proposed that work under complex lighting conditions~\cite{oghaz2015hybrid,lu2018color}.
We use the following skin detection rule based on the YCrCb space \cite{hsu2002face}, where $Y$ stands for luminance and $Cr,Cb$ stand for chrominance values.
\begin{align}\label{eq:skintonerule}
\text{skin} = \begin{cases}
\text{true}, & 90 \leq Cr \leq 115 \text{ and } 140 \leq Cb \leq 195 \\
\text{false}, &\text{otherwise}.
\end{cases}
\end{align}
We employ this rule for its simplicity and fairly good performance in skin detection under the favorable lighting conditions of the PPB* dataset.

\begin{figure}[ht]
\includegraphics[width=\columnwidth,trim={3cm 3cm 2cm 3cm},clip]{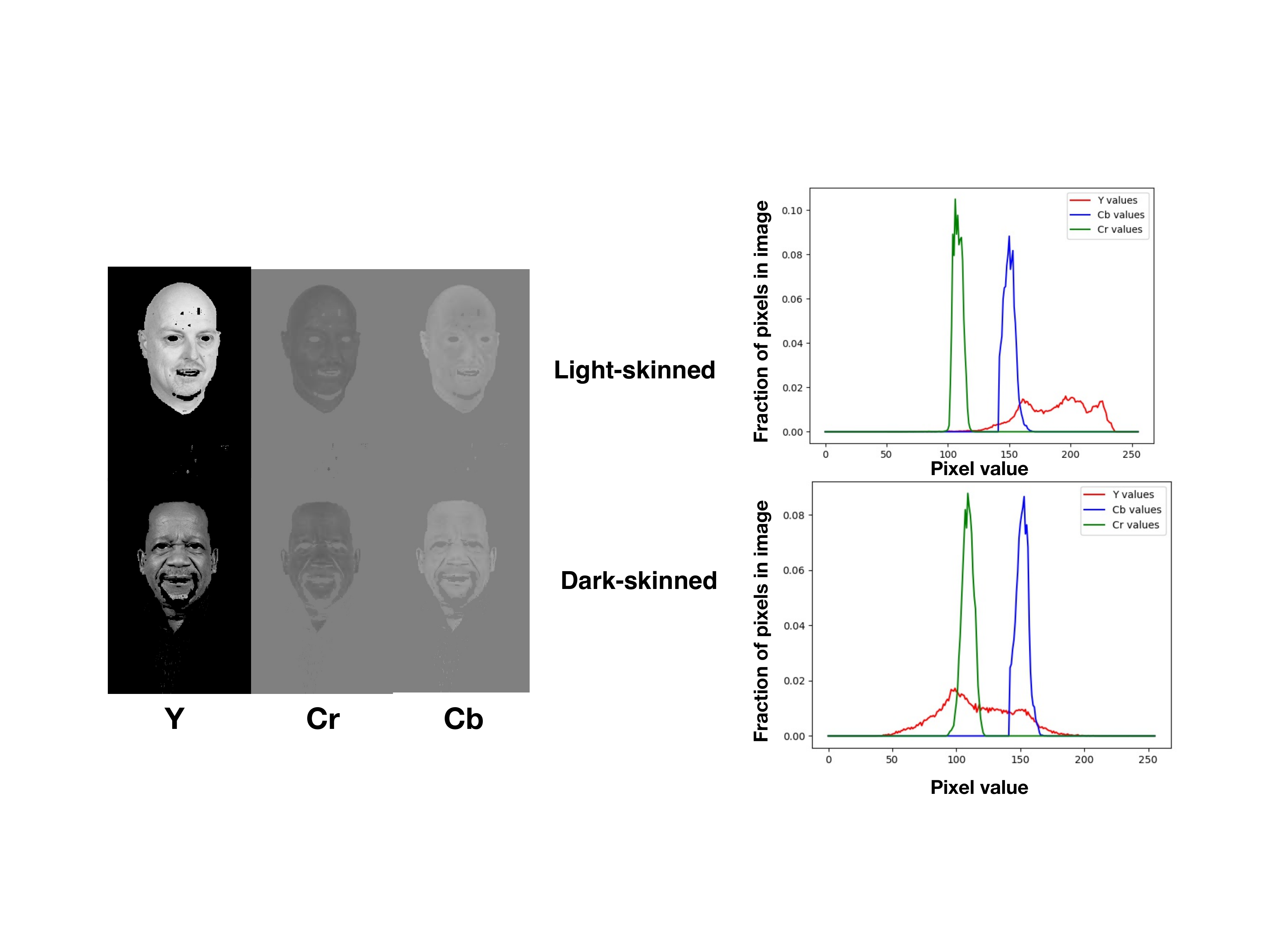}
\caption{Example of a light-skinned and dark-skinned image in the PPB* dataset. Observe that the Cr and Cb channels are similar across both images. Practically all variation in the skin type is captured in the Y component.}\label{fig:crcbeg}
\end{figure}

\begin{figure}[ht]
\includegraphics[width=\columnwidth,trim={4.5cm 5cm 7.5cm 5cm},clip]{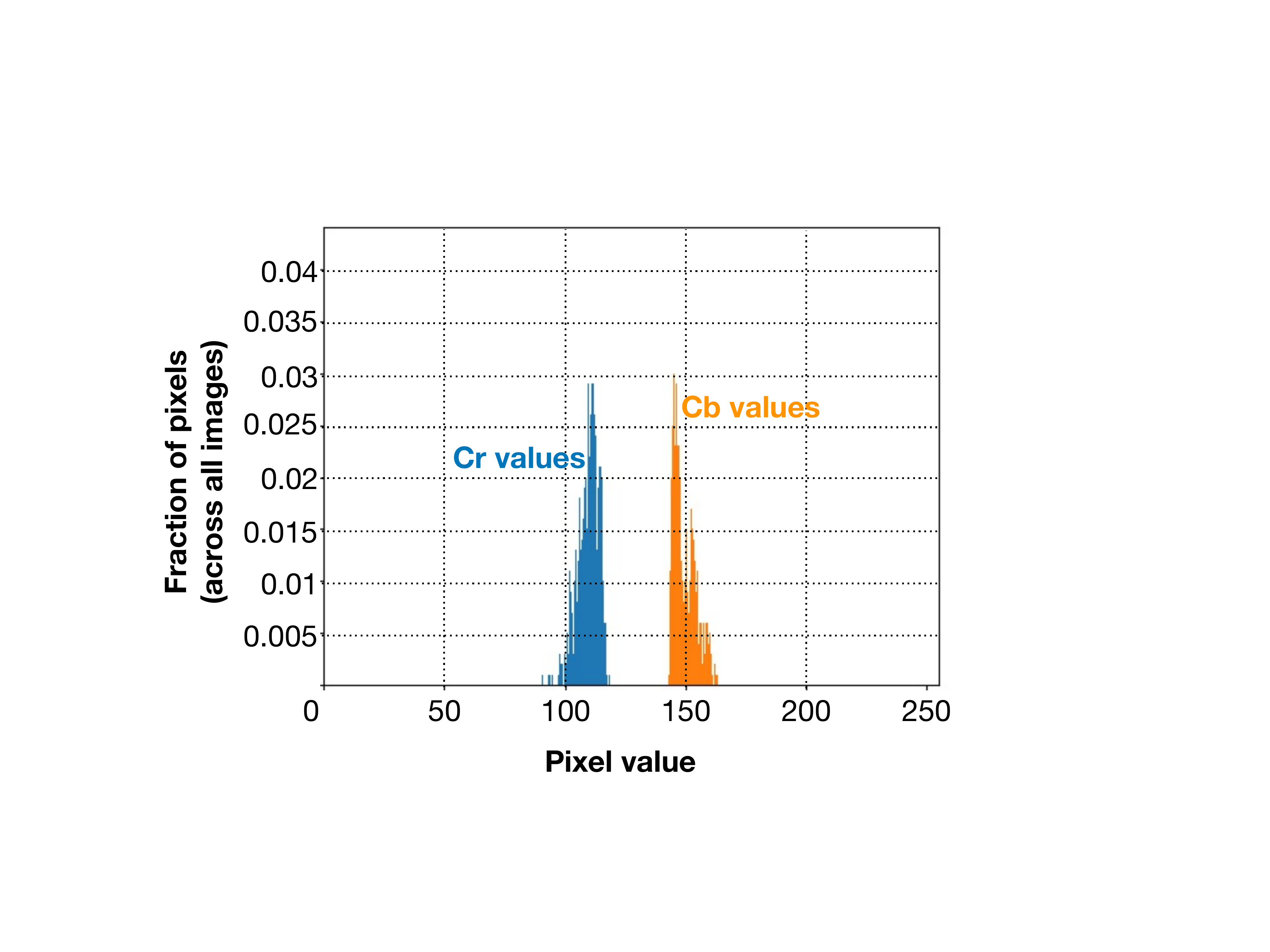}
\caption{Frequencies of Cr and Cb values across all skin type pixels across all images.}\label{fig:crcboverall}
\end{figure}

We also plot histograms of the YCrCb values of skin type pixels detected for each face image and observe that the $Cr$ and $Cb$ values fall into an even narrower range than described in \eqref{eq:skintonerule}.
As the illustration in Figure~\ref{fig:crcbeg} depicts, the chrominance values do not appear different for individuals with light or dark skin type.
More rigorously, Figure~\ref{fig:crcboverall} plots the histogram of $Cr$ and $Cb$ values across all $1204$ images in PPB*; we observe that the chrominance values are stable.
Practically all the variation in skin type is captured by the $Y$ channel alone.\footnote{We expect this phenomenon will hold for any face image with high resolution quality and uniform illumination.}

\subsubsection{Methods to change the skin type}

Based on the low-dimensional structure described in the previous subsection, we describe two rules that we employ to change the skin type of a face.
Both are carried out in the YCrCb color space.
We represent an image in RGB space by $I_{\mathsf{RGB}} \in [0,255]^{w \times h \times 3}$, and in YCrCb space by $I_{\mathsf{YCrCb}} \in [0,255]^{w \times h \times 3}$, where $(w,h)$ represents the width and height of the image.

\begin{figure*}[ht]
\includegraphics[height=2.3in]{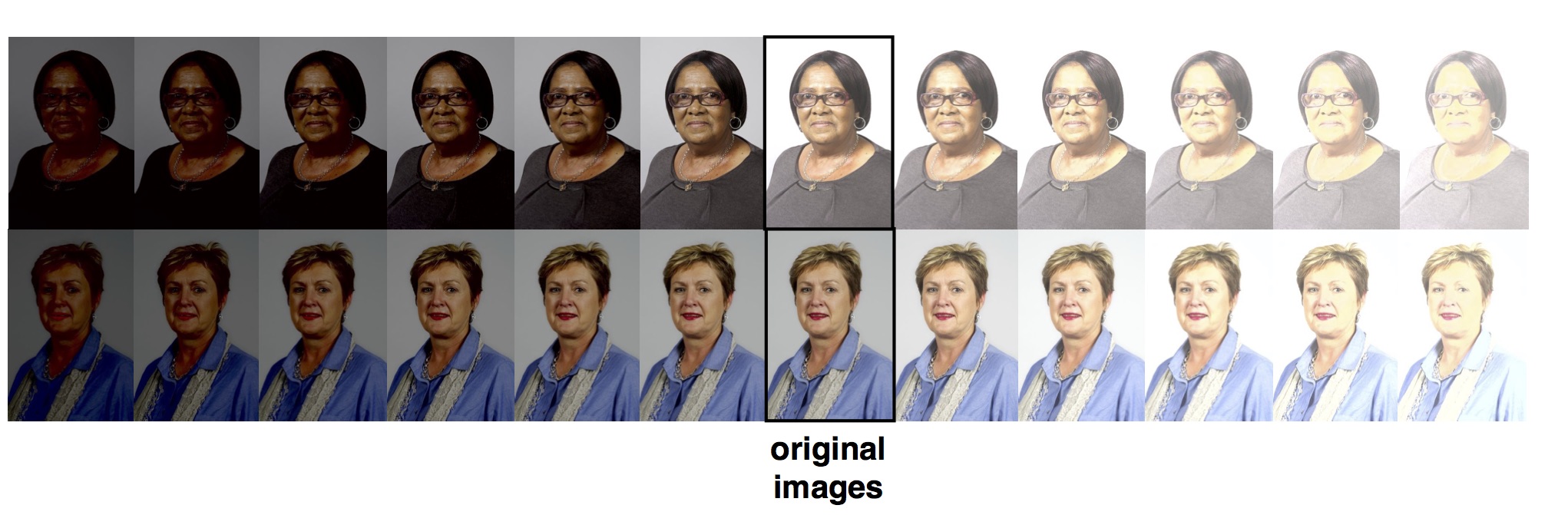}
\caption{Examples of light-skinned and dark-skinned faces whose luminance modes are shifted.}\label{fig:mode_visual}
\end{figure*}

\begin{procedure}[Luminance mode-shift]\label{def:modeshift}
We \textit{shift the skin type luminance mode} of an image in the following sequence of steps:
\begin{enumerate}
\item Determine $\text{old Y mode} = \text{Y mode}(\{I_{\mathsf{YCrCb}}(i,j) \in \text{ skin types} \})$.
\item Calculate the \textit{mode-shift-value} $\delta = \text{new Y mode} - \text{old Y mode}$.
\item Shift the luminance values, i.e. $I_{\mathsf{Y}} = I_{\mathsf{Y}} + \delta$.
\item Clip luminance values to $[0,255]$.
\end{enumerate}
\end{procedure}

\begin{figure*}[ht]
\includegraphics[width=\textwidth]{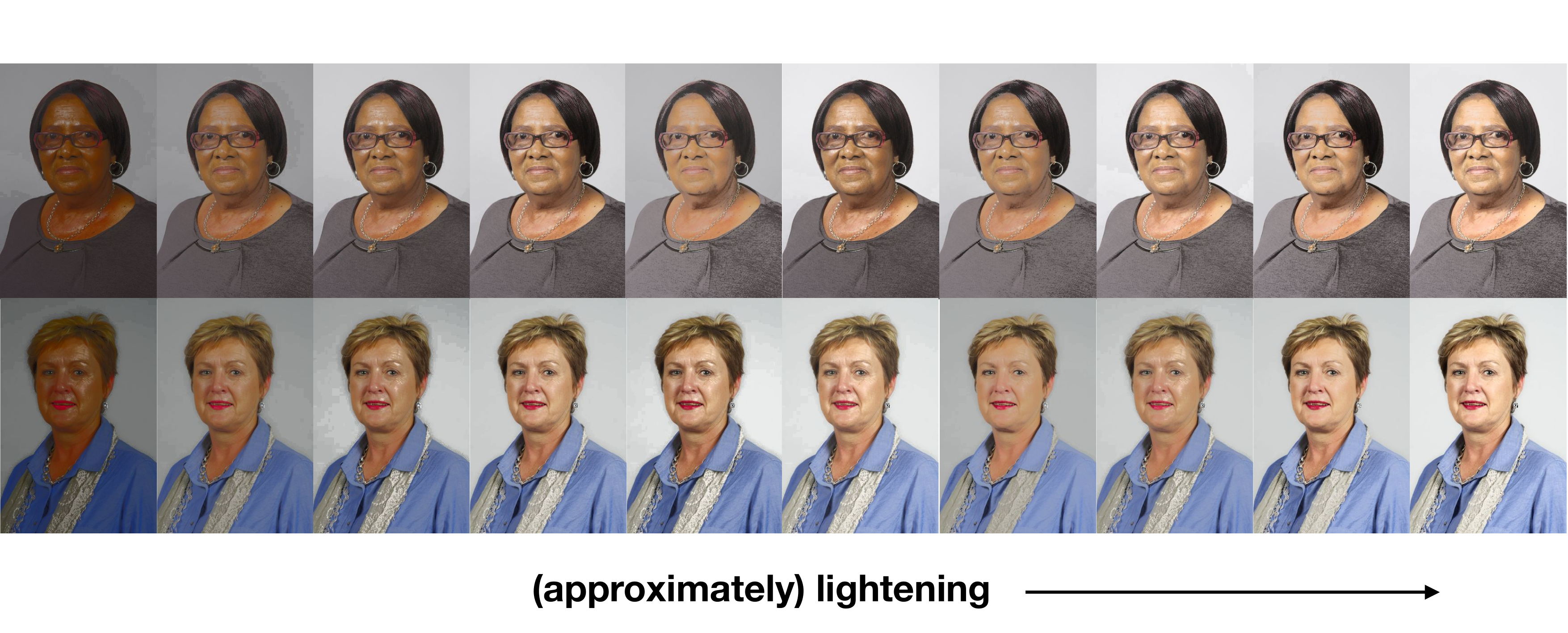}
\caption{Examples of light-skinned and dark-skinned faces that are optimally transported to new skin types, either darkened or lightened.}\label{fig:ot_visual}
\end{figure*}

Procedure~\ref{def:modeshift} is attractive for its simplicity and quick computation ($\mathcal{O}(1)$ time), but the results of skin type change according to luminance mode shift are not always visually attractive, as demonstrated in Figure~\ref{fig:mode_visual}.
Perhaps the luminance mode of skin type pixels is not sufficiently descriptive, and we would rather consider a transform between \textit{skin type histograms}.
Motivated by this, we next consider a skin type operation based on optimal transport, which has recently shown to be effective in color transfer in RGB space~\cite{ferradans2014regularized}.

\begin{procedure}[Optimal transport~\cite{ferradans2014regularized}]\label{def:ot}
This procedure takes as input a \textit{target skin type distribution over $Y$ values}.
We denote the skin type distribution of a grayscale image by $\mu(I)$ and the target skin type distribution by $\mu'_Y$. 
Then, the optimally transported image is defined as follows:
\begin{align*}
I^*_{\mathsf{Y}} := {\arg \min } ||I_{\mathsf{Y}} - I'_{\mathsf{Y}}||_2 \text { subject to } \mu_Y(I'_{\mathsf{Y}}) = \mu'_Y .
\end{align*}
\end{procedure}

Figure~\ref{fig:ot_visual} shows that the results of optimally transported skin type are visually more realistic.
However, the computational cost of using this operation is more; the optimal transport operation has complexity $\mathcal{O}((w \times h)^3)$.\footnote{The minimum size of images that we work with is $128 \times 128$, and in practice it takes $30$ seconds to a minute to optimally transport an image, compared to milliseconds to luminance mode shift an image.
For future work, we could utilize the computational reductions in computing the optimal transport using Sinkhorn regularization~\cite{cuturi2013sinkhorn}.}

\subsubsection{Results}

We consider the following ensemble of skin-type changes on the PPB* dataset:
\begin{enumerate}
\item Dark females/dark males: Evaluate the score on the original image. Evaluate the average new score on the set of lightened images.
\item Light females/light males: Evaluate the score on the original image. Evaluate the average new score on the set of darkened images.
\end{enumerate}
The set of darkened/lightened mode-shifted images represents all luminance-mode-shifts with negative/positive $\delta$.
Owing to the computational expense of optimal transport, we pick ten images on varying ends of the skin type spectrum. 

\begin{figure}
\centering
\begin{minipage}[b]{0.25\textwidth}
\includegraphics[height=1.3in,trim={2cm 1.5cm 2cm 2cm},clip]{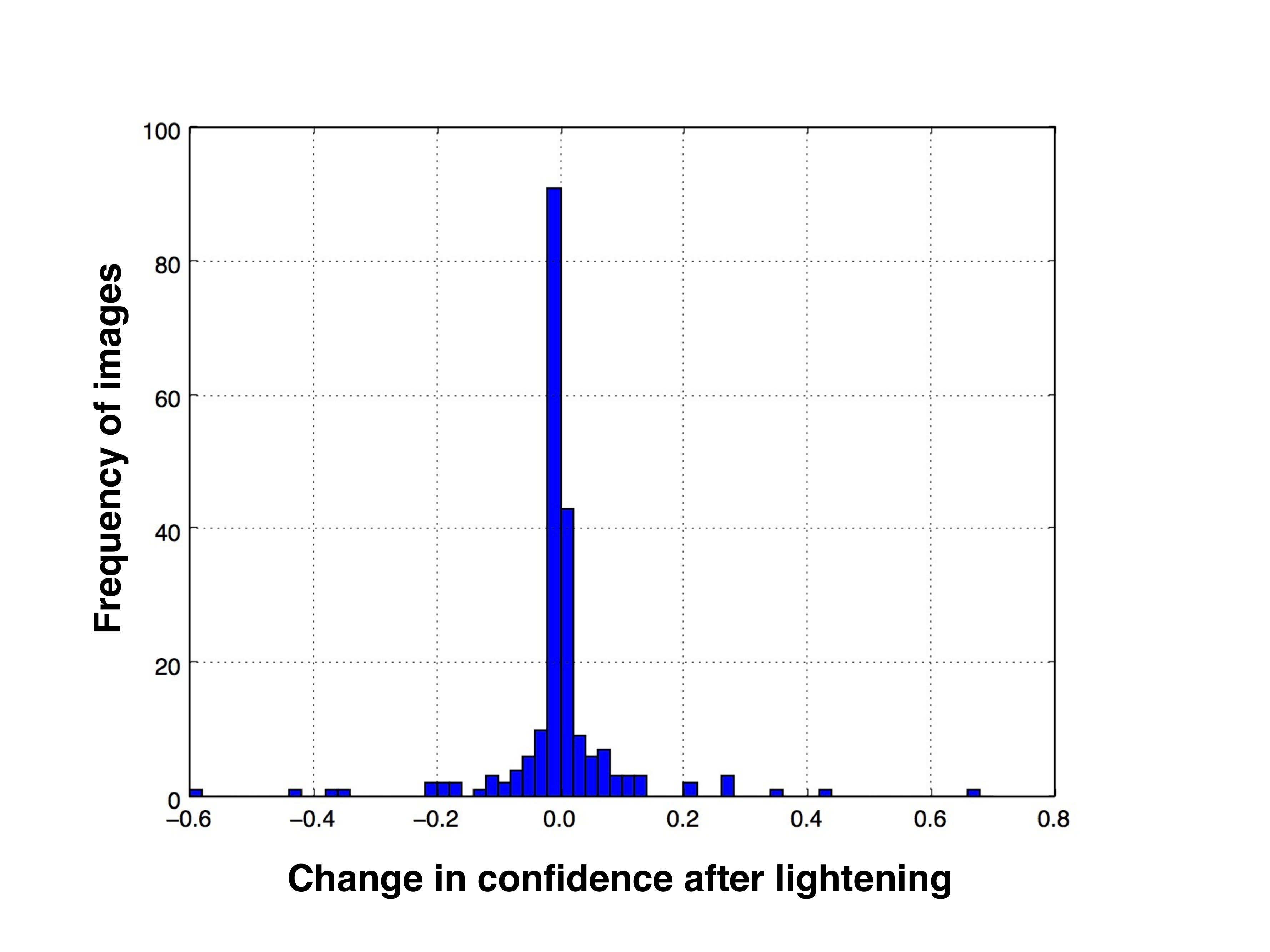}\\
\subcaption{Luminance-mode-shift.}\label{fig:histogram_DF_modeN}
\end{minipage}%
\begin{minipage}[b]{0.25\textwidth}
\includegraphics[height=1.3in,trim={2cm 1.5cm 2cm 2cm},clip]{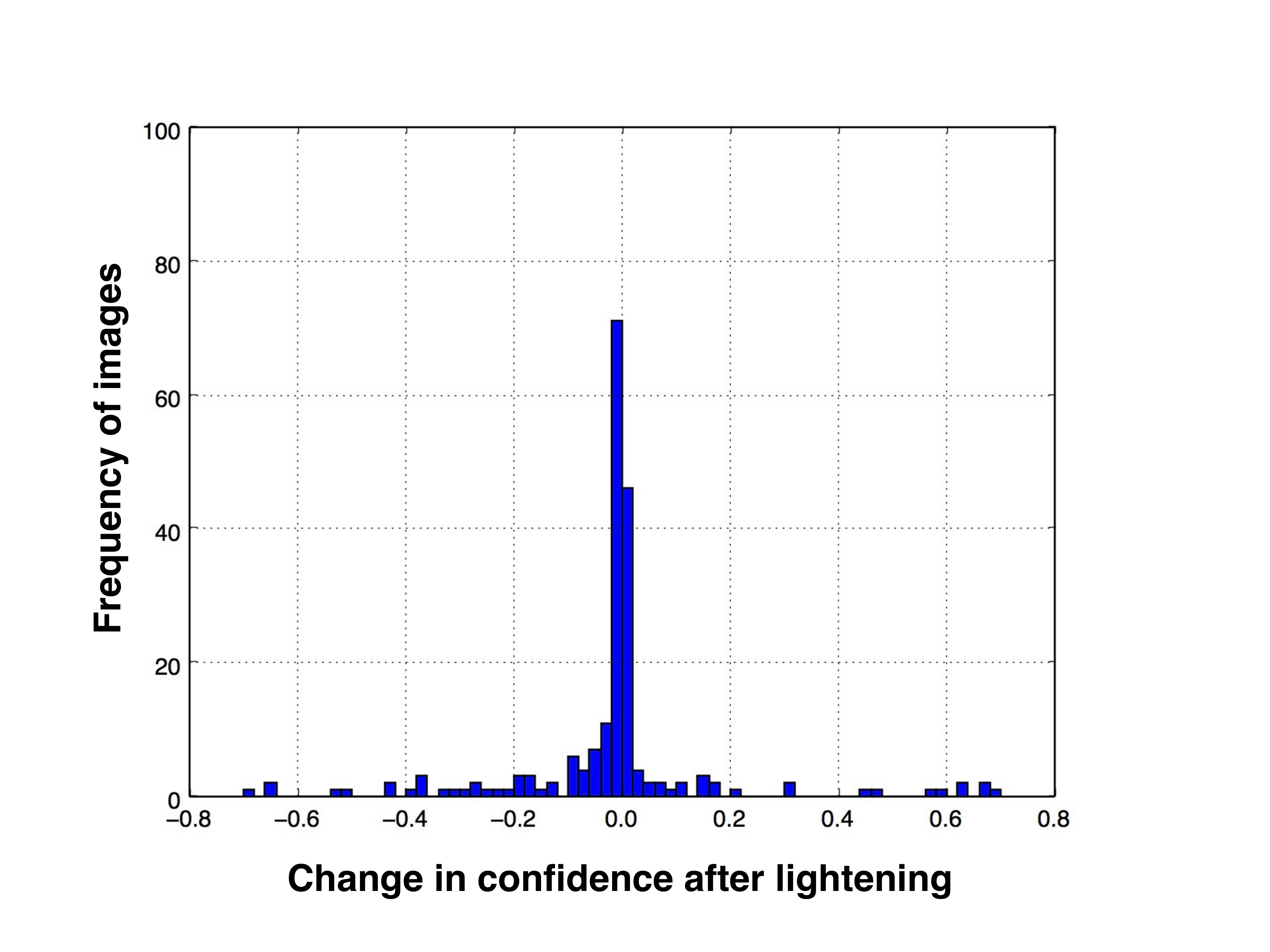}\\
\subcaption{Optimal transport.}\label{fig:histogram_DF_OT}
\end{minipage}%
\caption{Histograms of differences in scores of dark females in PPB* dataset after lightening the skin type.}\label{fig:histogram_DF}
\end{figure}

\begin{figure}
\centering
\begin{minipage}[b]{0.25\textwidth}
\includegraphics[height=1.3in,trim={5cm 4.5cm 5cm 5cm},clip]{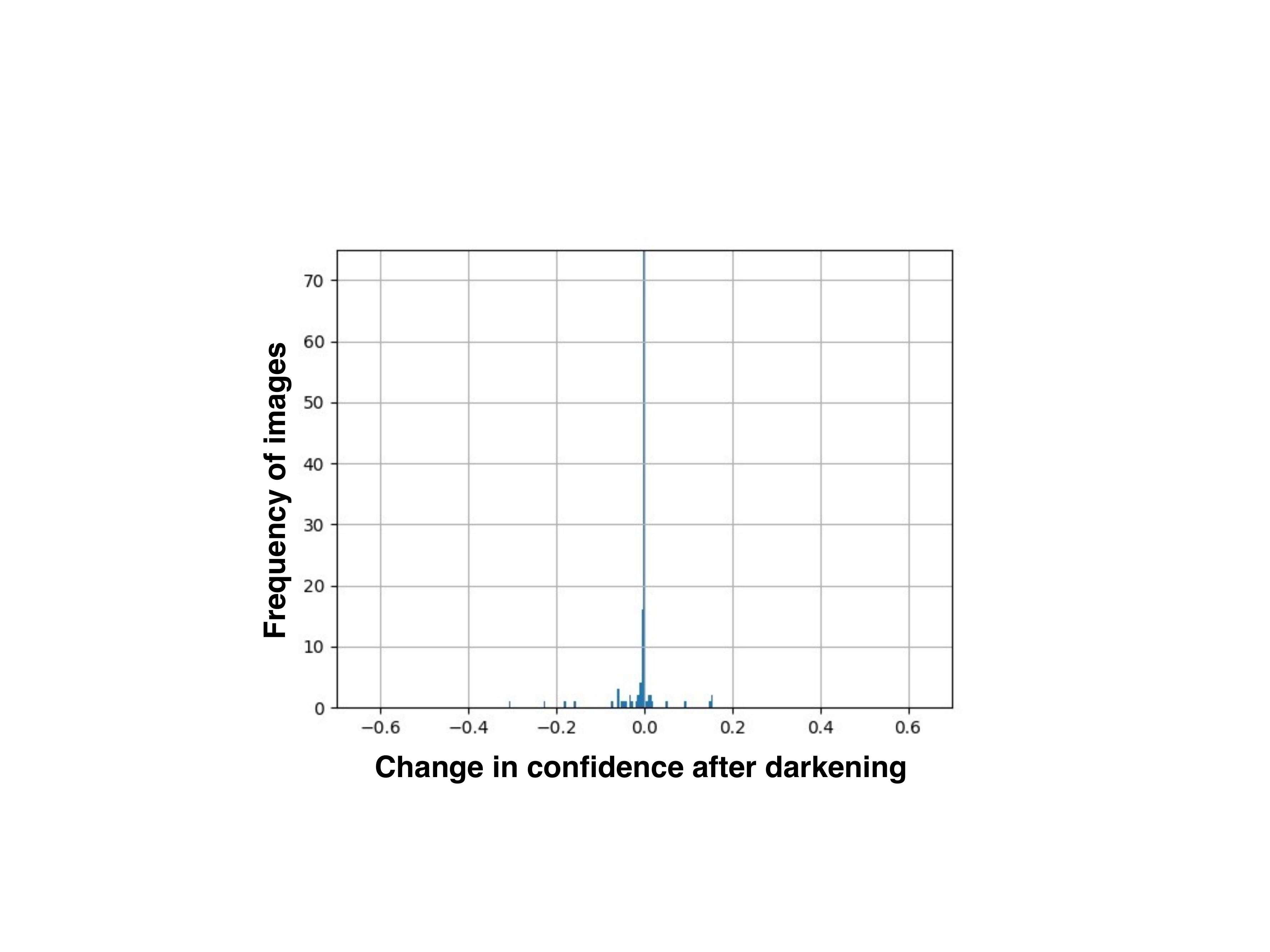}\\
\subcaption{Luminance-mode-shift.}\label{fig:histogram_LF_modeN}
\end{minipage}%
\begin{minipage}[b]{0.25\textwidth}
\includegraphics[height=1.3in,trim={5cm 4.5cm 5cm 5cm},clip]{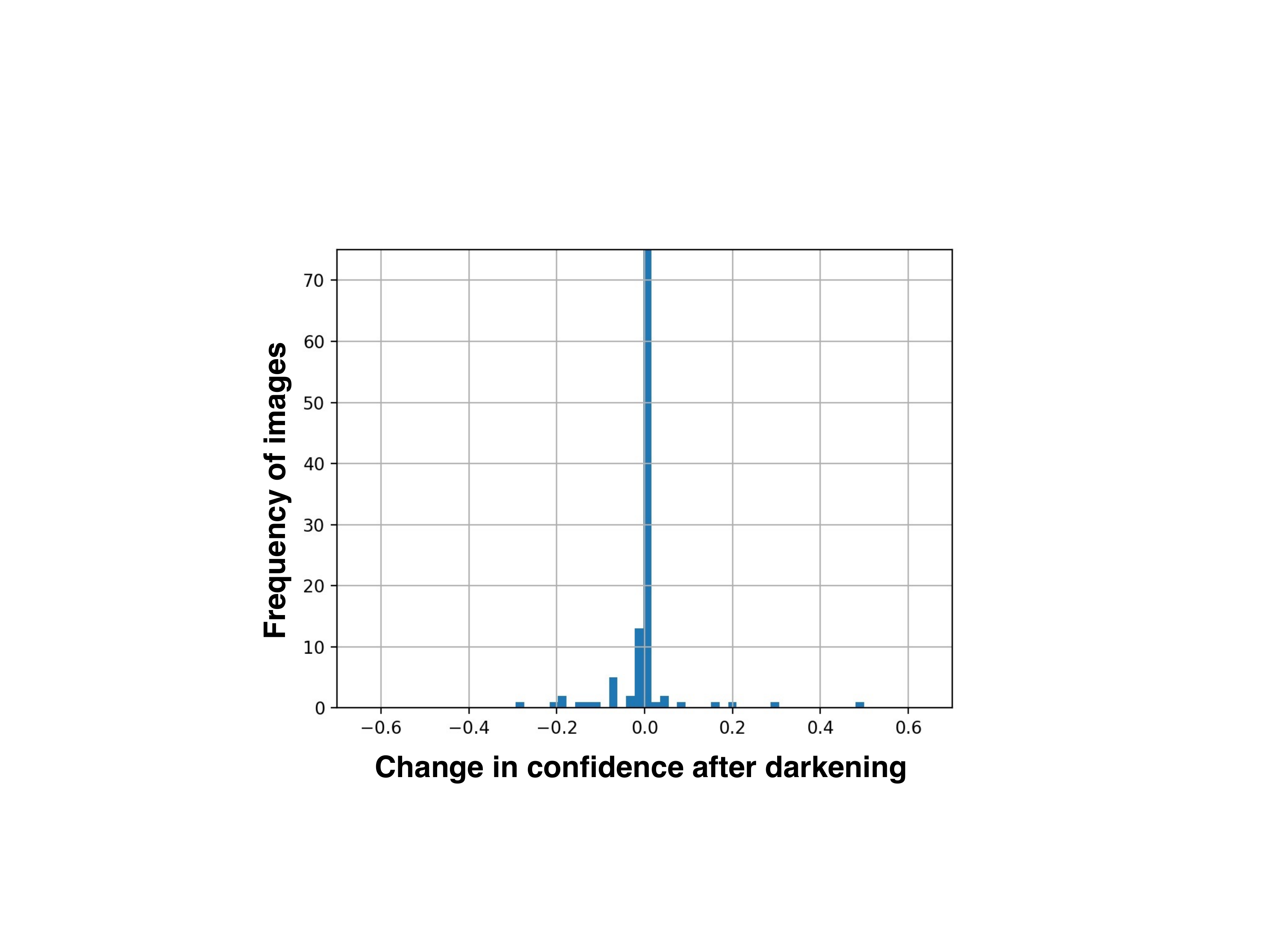}\\
\subcaption{Optimal transport.}\label{fig:histogram_LF_OT}
\end{minipage}%
\caption{Histograms of differences in scores of light females in PPB* dataset after darkening the skin type.}\label{fig:histogram_LF}
\end{figure}

\begin{figure}
\begin{minipage}[b]{0.25\textwidth}
\centering
\includegraphics[height=1.3in,trim={2cm 1.5cm 2cm 2cm},clip]{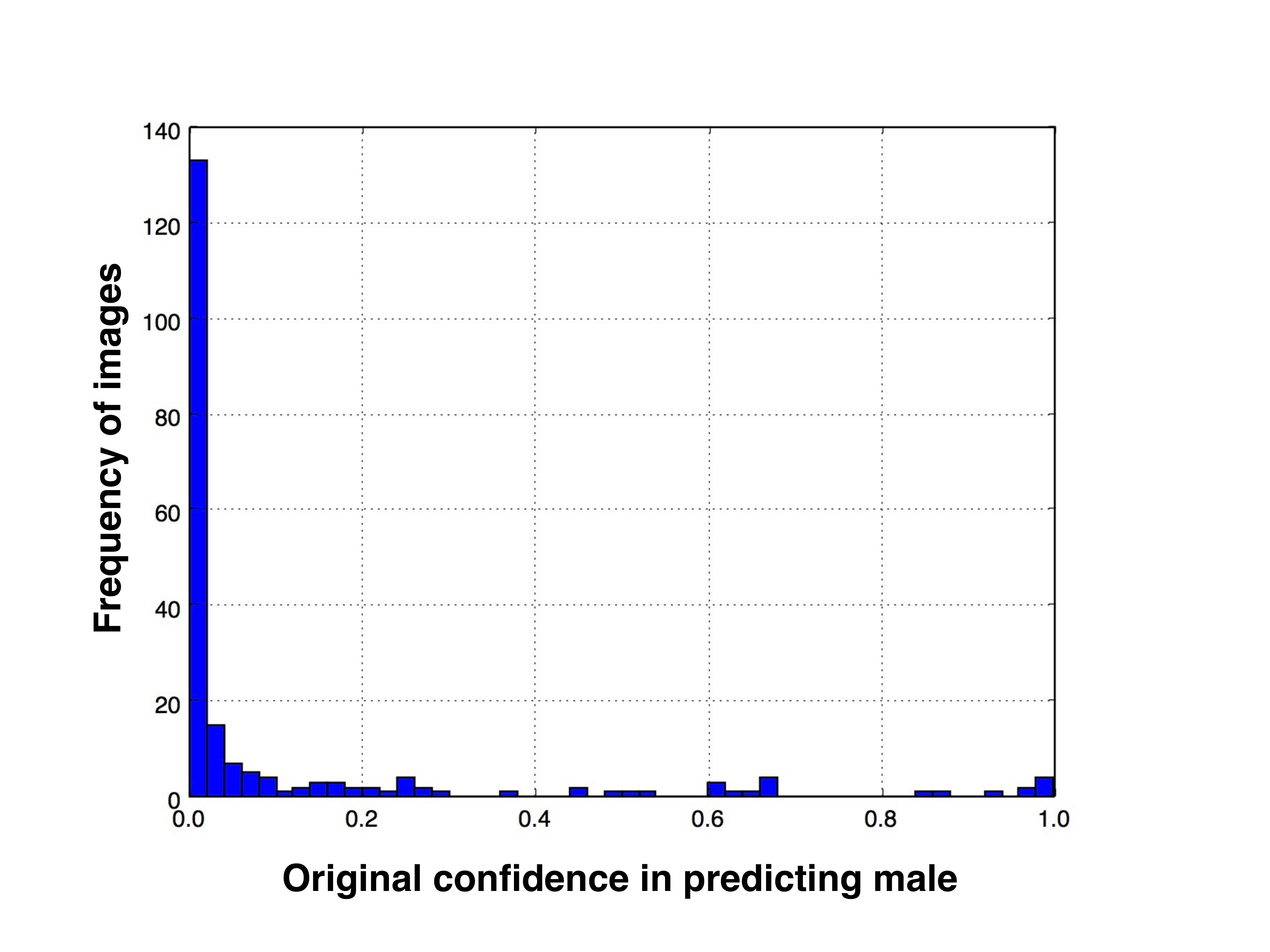}\\
\subcaption{Scores.}\label{fig:confidences_DF}
\end{minipage}
\\
\begin{minipage}[b]{0.25\textwidth}
\includegraphics[height=1.3in,trim={2cm 1.3cm 2cm 2cm},clip]{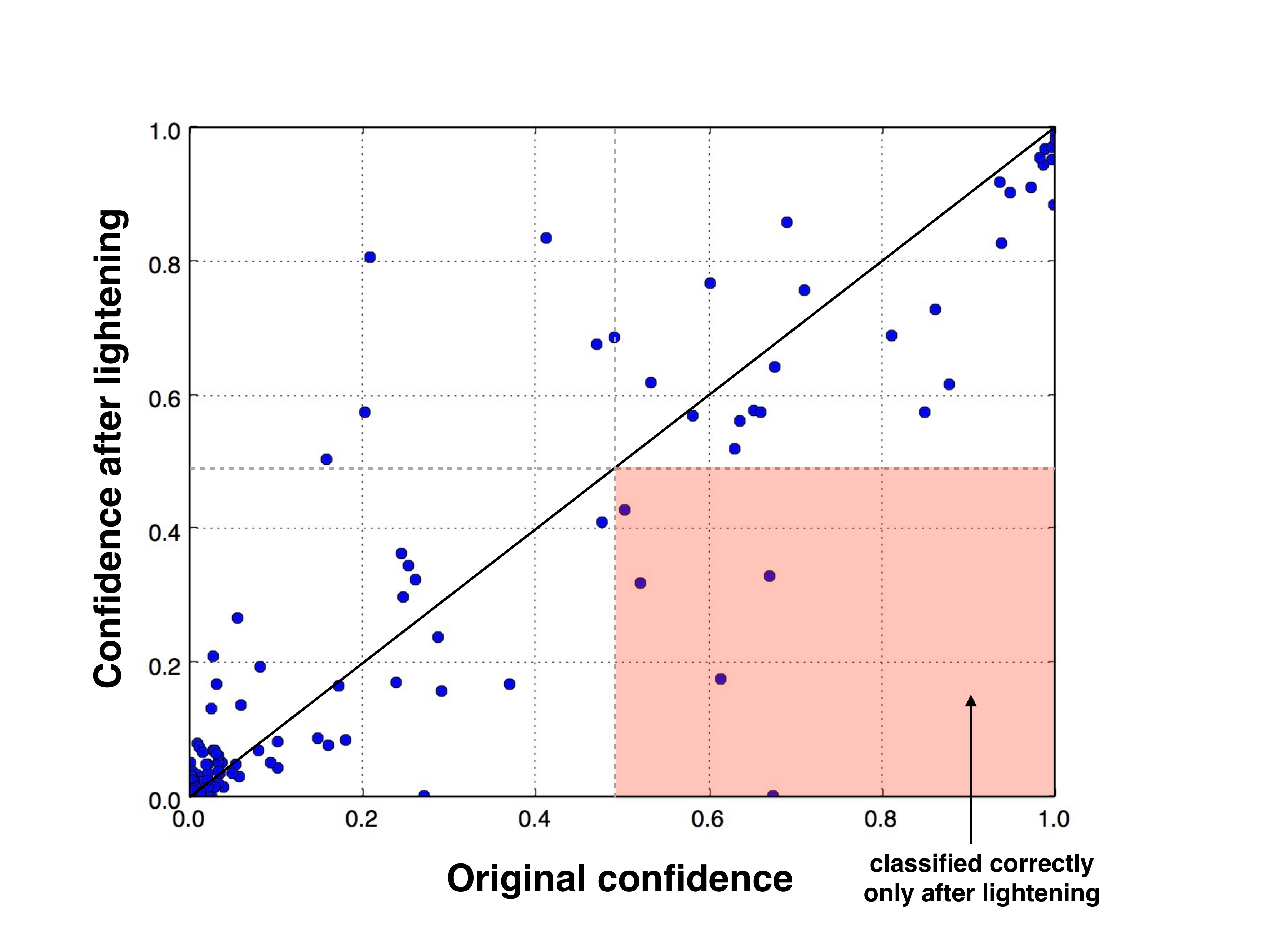}\\
\subcaption{Mode-shift.}\label{fig:scatterplot_DF_modeN}
\end{minipage}%
\begin{minipage}[b]{0.25\textwidth}
\includegraphics[height=1.3in,trim={2cm 1.3cm 2cm 2cm},clip]{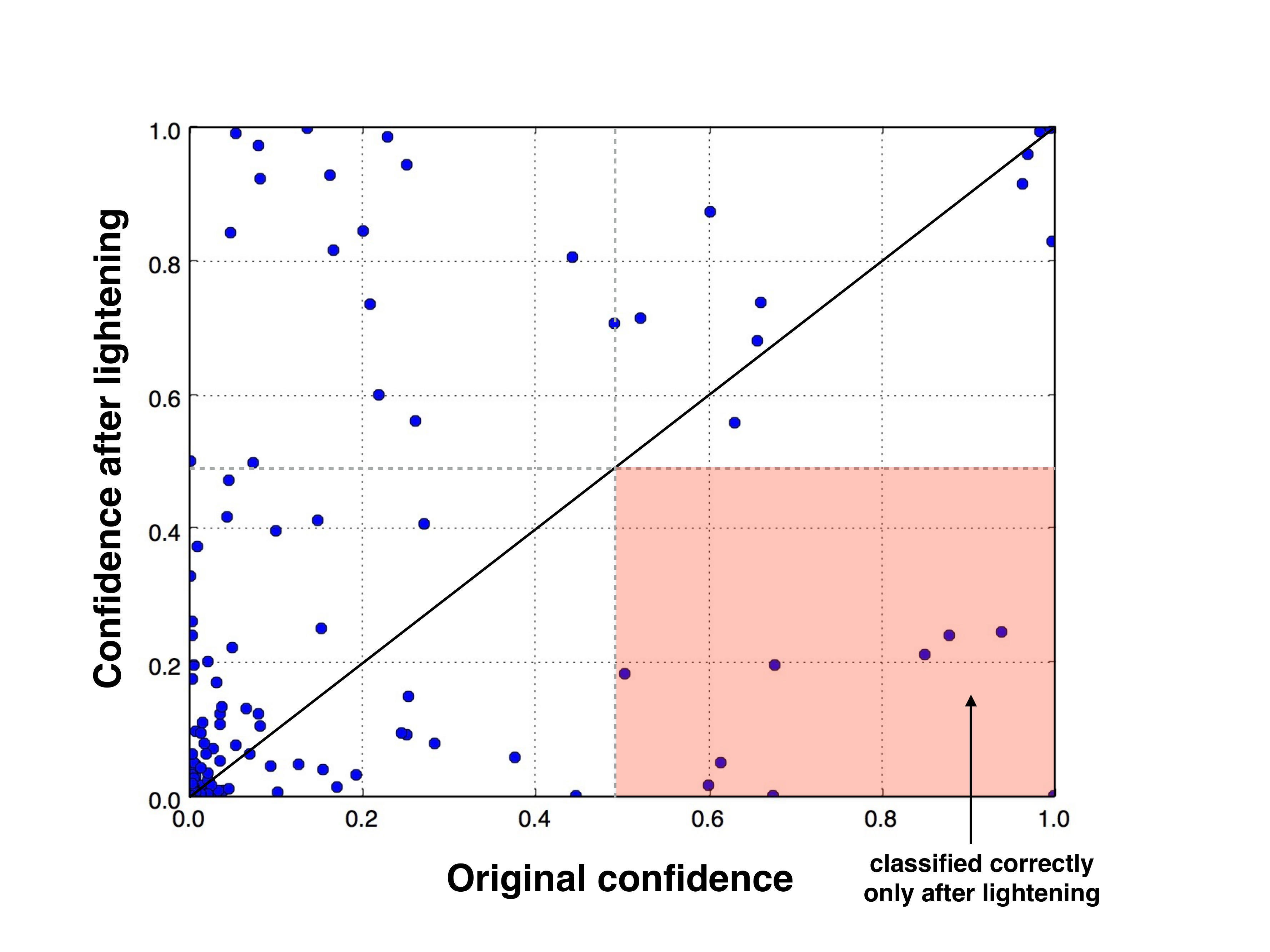}\\
\subcaption{OT.}\label{fig:scatterplot_DF_OT}
\end{minipage}%
\caption{Scatterplots of original prediction vs prediction after lightening for dark females. Shaded region represents dark females correctly classified after lightening.}\label{fig:scatterplots_DF}
\end{figure}

Figure~\ref{fig:histogram_DF} shows the distribution of affected differences in prediction on lightening the set of dark females in the PPB dataset, either using mode-shift (Figure~\ref{fig:histogram_DF_modeN}) or optimal transport (Figure~\ref{fig:histogram_DF_OT}).\footnote{The quality of the experiment itself is better with the optimal transport method as the lightened images are more realistic, but owing to computational complexity of optimal transport, we also have fewer lightened samples to average over.
On the other hand, the mode-shift operation generates images that are not as realistic, but the experiment itself is statistically more robust as we can quickly generate many lightened samples.
Thus, observing similar conclusions for the two methods strengthens our result.
}
We observe that most images' scores do not change meaningfully after lightening/darkening.
In the case of dark females, $86.6\%$ of the images' scores do not change by more than $0.1$ on lightening using mode-shift.
$76.6\%$ of the images' scores do not change by more than $0.1$ on lightening using optimal transport.
In the case of light females, $96.3\%$ of the images' scores do not change by more than $0.1$ on darkening using mode-shift.
$92.1\%$ of the images' scores do not change by more than $0.1$ on darkening using optimal transport.
We conducted one-sample $t$-tests to test the null hypothesis that the mean of differences in scores is equal to $0$.
The results in terms of the $95\%$ confidence intervals are presented in Table~\ref{tab:stat_tests}.

\begin{figure}[ht]
\begin{minipage}[b]{0.25\textwidth}
\centering
\includegraphics[height=1.3in,trim={5cm 2cm 5cm 5cm}]{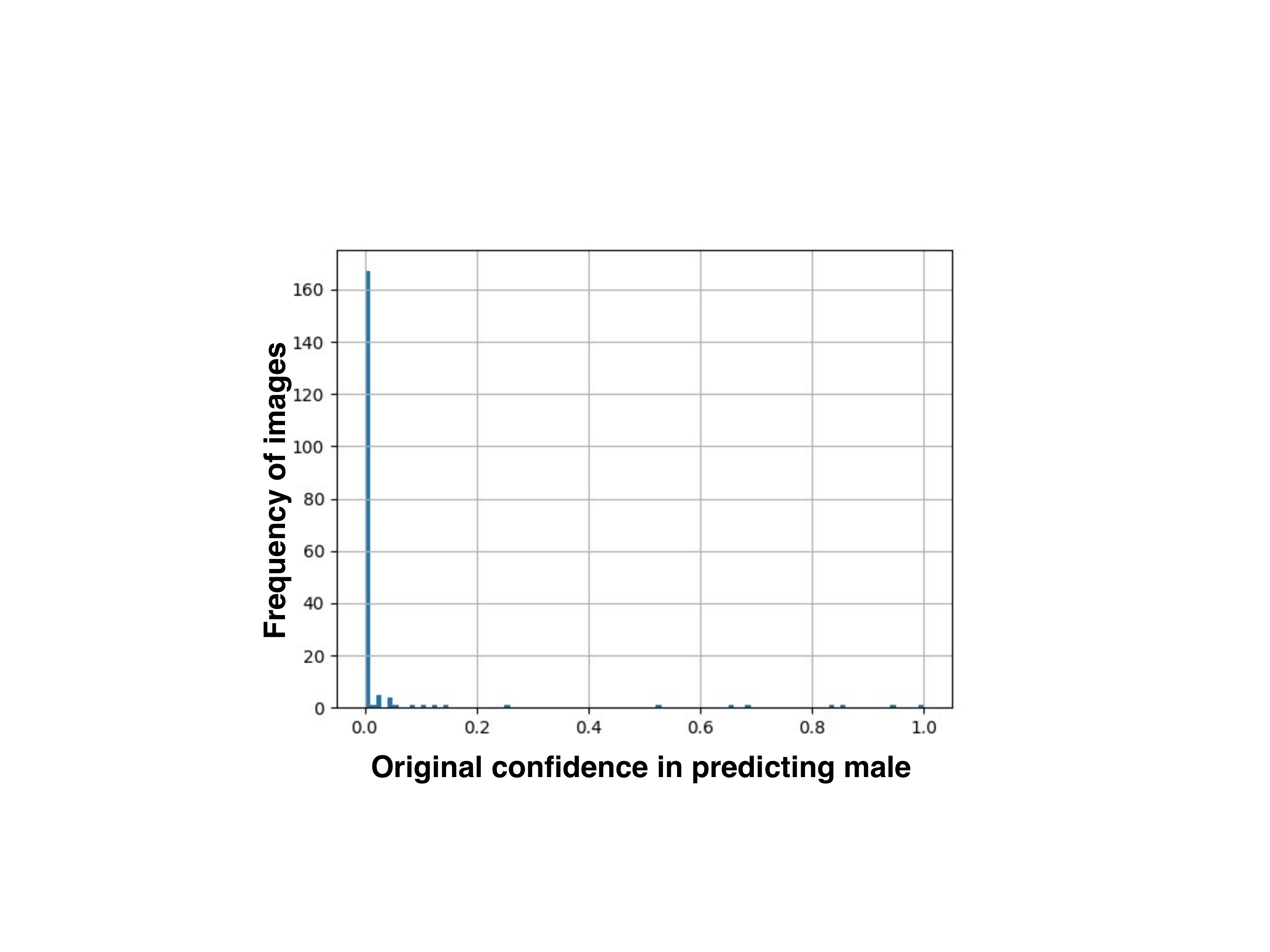}\\
\subcaption{Scores.}\label{fig:confidences_LF}
\end{minipage}%
\\
\begin{minipage}[b]{0.25\textwidth}
\includegraphics[height=1.3in,trim={5cm 4.5cm 5cm 5cm},clip]{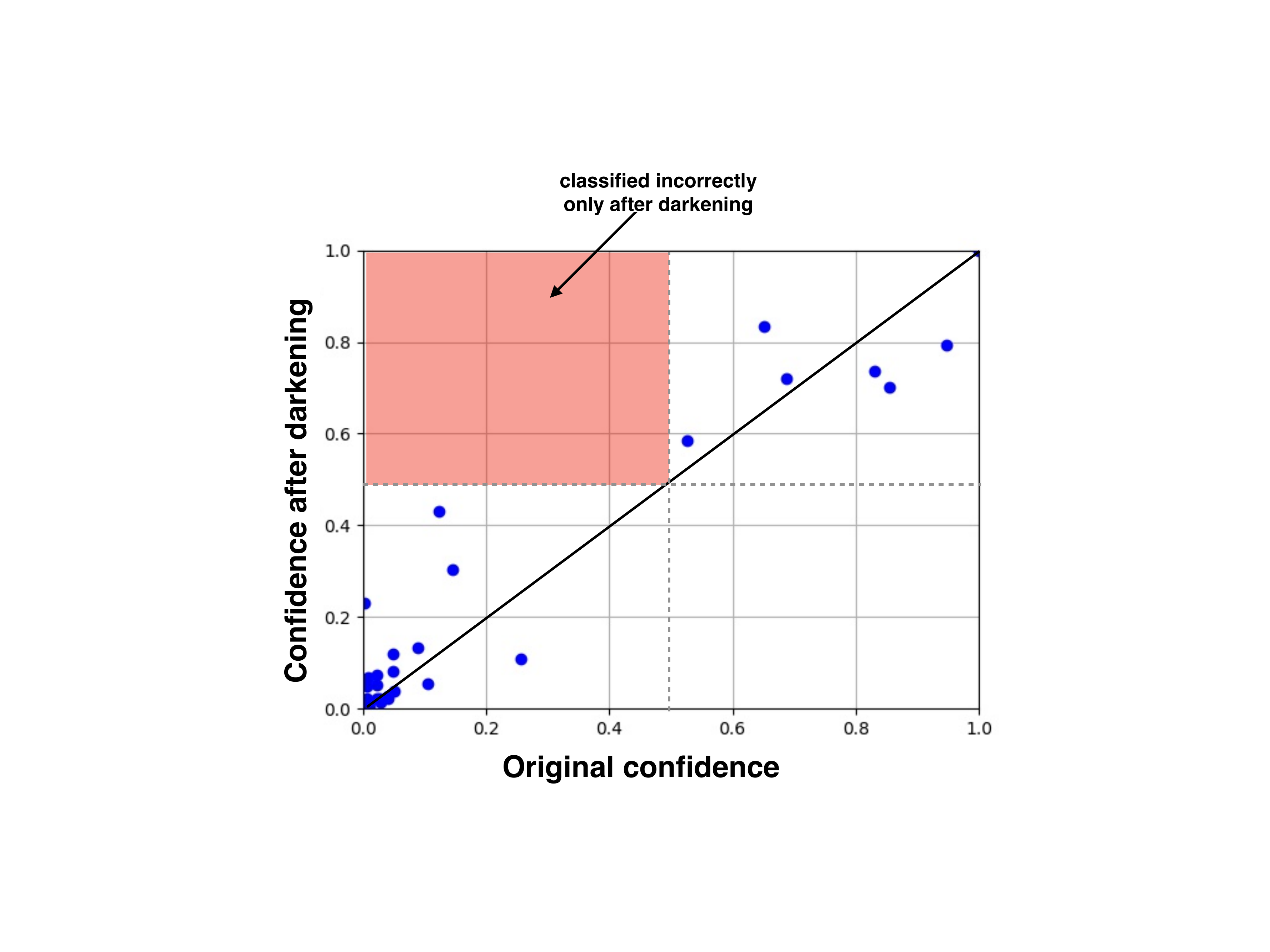}\\
\subcaption{Mode-shift.}\label{fig:scatterplot_LF_modeN}
\end{minipage}%
\begin{minipage}[b]{0.25\textwidth}
\includegraphics[height=1.3in,trim={5cm 4.5cm 5cm 5cm},clip]{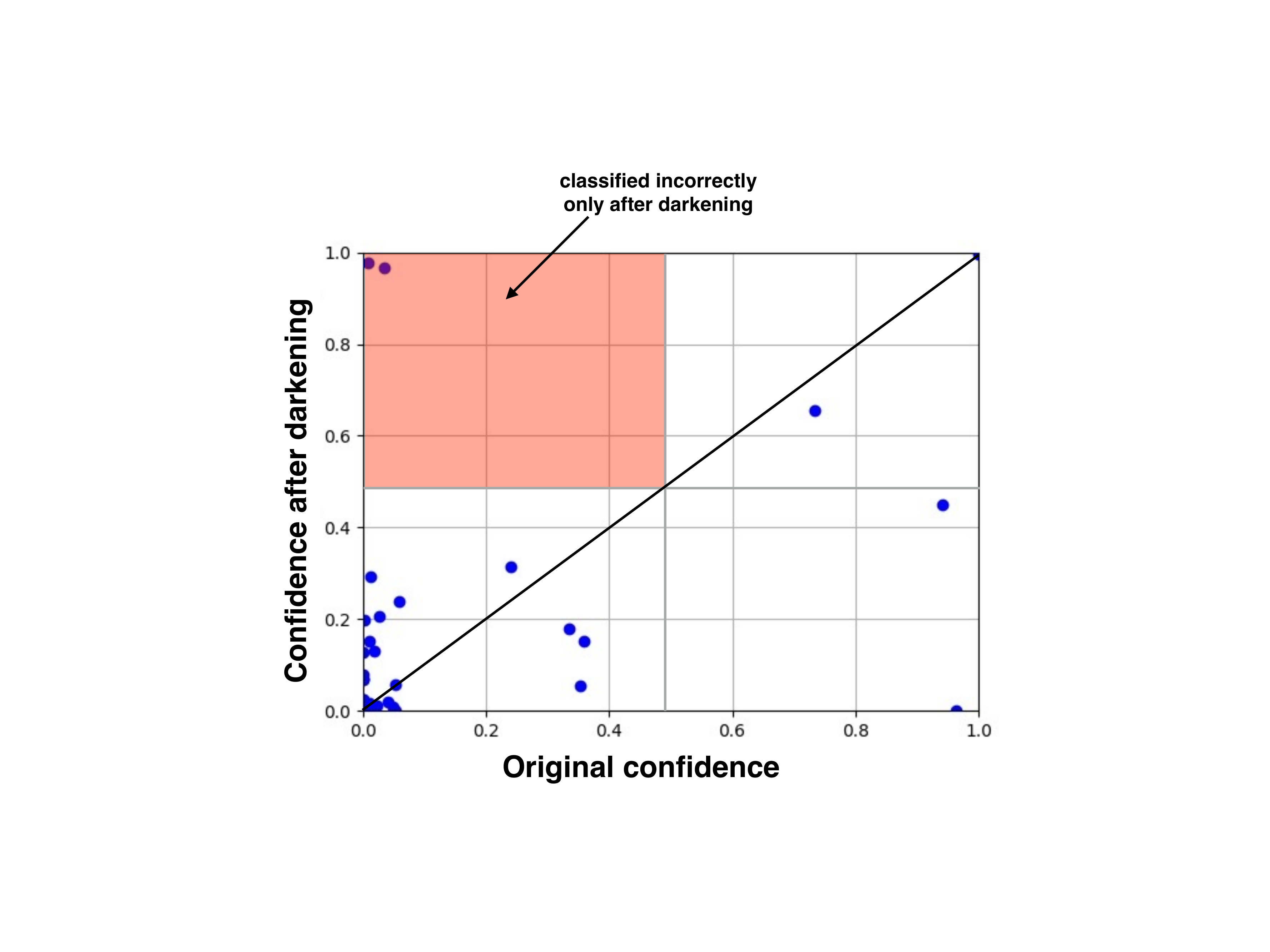}\\
\subcaption{OT.}\label{fig:scatterplot_LF_OT}
\end{minipage}%
\caption{Scatterplots of original prediction vs prediction after darkening for light females. The shaded region represents light females that would be correctly classified after darkening.}\label{fig:scatterplots_LF}
\end{figure}

\begin{table}
\centering
\caption{Results of one sample t-test on mean of differences in scores with respect to $0$ after skin type change.}
  \label{tab:stat_tests}
  \begin{tabular}{cccc}
    \toprule
    Category & $95\%$ confidence interval  \\
    \midrule
    DF, mode-shift &  $[-0.013,0.015]$  \\
    \midrule
    DF, OT &  $[-0.071,-0.003]$ \\
    \midrule
    LF, mode-shift &  $[-0.010,0.001]$\\
    \midrule
    LF, OT &  $[-0.035, 0.016]$
\\
    \bottomrule
  \end{tabular}
\end{table}

Figures~\ref{fig:scatterplots_DF} and~\ref{fig:scatterplots_LF} shed insight into the \textit{relative difference in predictions}, which also matters --- in particular, we may care about the fraction of images whose average classification decision would change after lightening/darkening.
In the scatterplots of original score vs.\ score after change in skin type (Figures~\ref{fig:scatterplot_DF_modeN},~\ref{fig:scatterplot_DF_OT},~\ref{fig:scatterplot_LF_modeN} and~\ref{fig:scatterplot_LF_OT}), we highlight the points that fall in the red-shaded region as representing dark females that are correctly classified \textit{only after lightening}, or light females that are incorrectly classified \textit{only after darkening}.
Very few images fall into these categories: $5$ and $9$ dark females (out of $212$) are correctly classified only after lightening using mode-shift and optimal transport respectively.
The effect of darkening is even less pronounced for light females -- after mode-shift and optimal transport, $0$ and $2$ females (out of $296$) respectively become incorrectly classified.
Looking at the distribution on original scores of dark and light females (Figures~\ref{fig:histogram_DF} and~\ref{fig:histogram_LF}), we see that almost all light females are classified as female with extremely high score, and almost all dark females are classified as either female or male with extremely high score.
The dark females that are classified as male with extremely high score, say above $0.9$, do not change significantly in score or classification decision on lightening.

All of these results, together, lead us to conclude that \textit{the skin type by itself has a minimal effect on classification decisions}.

\subsection{The potential influence of hair length}

If the skin type is not an underrepresented facial feature that matters, then what is?
We visually observed that most of the dark females that were misclassified as male by the Watson classifier had short hair.
We manually labeled all the dark and light females in the PPB* dataset as either short or long-haired, and considered the intersectional performance of the Watson classifier on females across skin type and hair length.\footnote{We did not consider a similar split across males, because all males in the PPB* dataset are short-haired.}
The results are presented in Table~\ref{tab:accuracies_hairlength}.
We notice a meaningful split in performance across hair length, especially for dark females: the classification accuracy on dark females with short hair is only $75\%$, as opposed to a much higher $92\%$ on dark females with long hair.
While the difference is less pronounced for light females, the accuracy on short-haired light females is also lower at $90\%$.
We also observed a relatively higher proportion of short-haired dark females in the PPB* dataset:\footnote{We are not making this claim for the general population of light-skinned and dark-skinned females, only the ones in the PPB* dataset.} $72\%$ of the dark females were observed to be short-haired, as opposed to $46\%$ of the light females.
Looking at purely misclassified \textit{dark females}, $29$ out of $33$ were short-haired!

\begin{table}
\centering
  \caption{Accuracy of Watson classifier on females intersected across skin type and hair length.}
  \label{tab:accuracies_hairlength}
  \begin{tabular}{ccc}
    \toprule
    & Dark-skinned & Light-skinned \\
    \midrule
    Short-haired & $75\%$ & $90\%$ \\ \midrule
    Long-haired & $92\%$ & $99.5\%$ \\   
    \bottomrule
  \end{tabular}
\end{table}

A hypothesis to explain the unequal performance displayed above is that the neural networks latch on to hair length as a significant predictor for gender.
It is well-known in human visual perception that certain hairstyles (including male facial hair) are convincingly attributed to respective (binary) genders \cite{brown1993gives}.
Such simplistic explanations can lead not only to racial biases but also confirmations of \textit{gender stereotypes}.

\begin{figure}[t]
\includegraphics[width=\columnwidth]{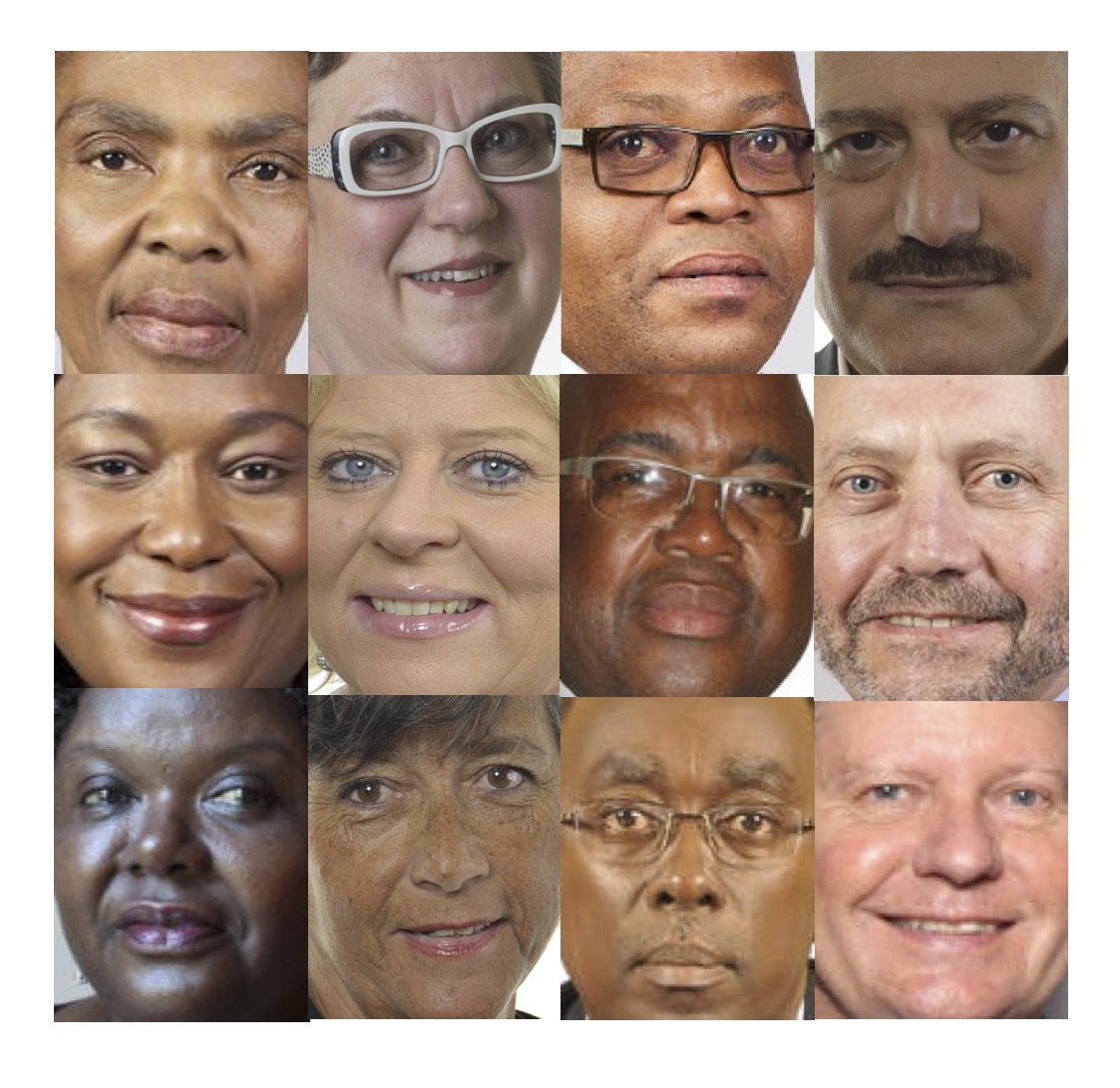}
\caption{Examples of images in PPB* dataset, only-face cropped.}\label{fig:faces_onlyface}
\end{figure}

We do not replicate the stability experiment for hair patterns because it is challenging to develop methodology to change a face's hairstyle while keeping all other attributes unchanged.\footnote{While some GANs are able to do this in theory~\cite{choi2017stargan}, the visual results are unconvincing.
Moreover, these will not be able to adapt to hairstyles of underrepresented ethnicities.}
We can do a different sort of experiment, however --- we can consider the performance of state-of-the-art gender classifiers \textit{that completely ignore information about the hair}.
This is achieved by using \textit{cropped facial images} as input obtained via standard face detectors in computer vision~\cite{viola2001rapid}.
By definition, these input images will contain only facial information, and not information about hair patterns (Figure~\ref{fig:faces_onlyface} contains some example images).

We achieve intersectional accuracies on the PPB* dataset reported in Table~\ref{tab:accuracies_SVM} and Table \ref{tab:accuracies_SVM_new} using SVMs with Watson deep face features and ResNet-50 deep face features with different kernel parameters on only-face cropped images, .
We observe that the accuracies are greater than $90\%$ for males and relatively high for light females ($80\%$), but only $66\%$ for dark females with the Watson features. Similarly, but at a higher overall accuracy level, the accuracies are greater than $95\%$ for males and light-skinned females, but only around $80\%$ for dark-skinned females.  
Thus, the unequal performance persists even with a state-of-the-art gender classifier that has no information about hair, suggesting that there are other underrepresented features in the face that result in differential performance. Seeing the same result with different feature extraction methods trained on different datasets only strengthens the conclusion. 

\begin{table}
\centering
  \caption{Accuracy of SVM trained on Watson deep face features intersected across skin type and gender.}
  \label{tab:accuracies_SVM}
  \begin{tabular}{ccc}
    \toprule
    & Female & Male \\
    \midrule
    Dark-skinned & $66.3\%$ & $91.5\%$ \\
    \midrule
    Light-skinned & $80.6\%$ & $96.9\%$ \\ 
    \bottomrule
  \end{tabular}
\end{table}
\begin{table}
    \centering
    \begin{tabular}{ccc}
	\toprule
         & Female & Male \\
         \midrule
       Dark-skinned  & $82.46/75.35\%$  & $95.53/97.59\%$ \\
       \midrule
       Light-skinned & $97.56/95.47\%$ & $97.19/100\%$ \\
	\bottomrule
    \end{tabular}
    \caption{Accuracy of SVM (for RBF kernel parameters $10^{-4}/10^{-5}$) trained on ResNet-50 deep face features intersected across skin type and gender.}
    \label{tab:accuracies_SVM_new}
\end{table}

\subsection{Sufficient facial features}
\begin{figure}[ht]
\centering
\begin{minipage}[b]{0.4\textwidth}
\includegraphics[width=0.95\columnwidth]{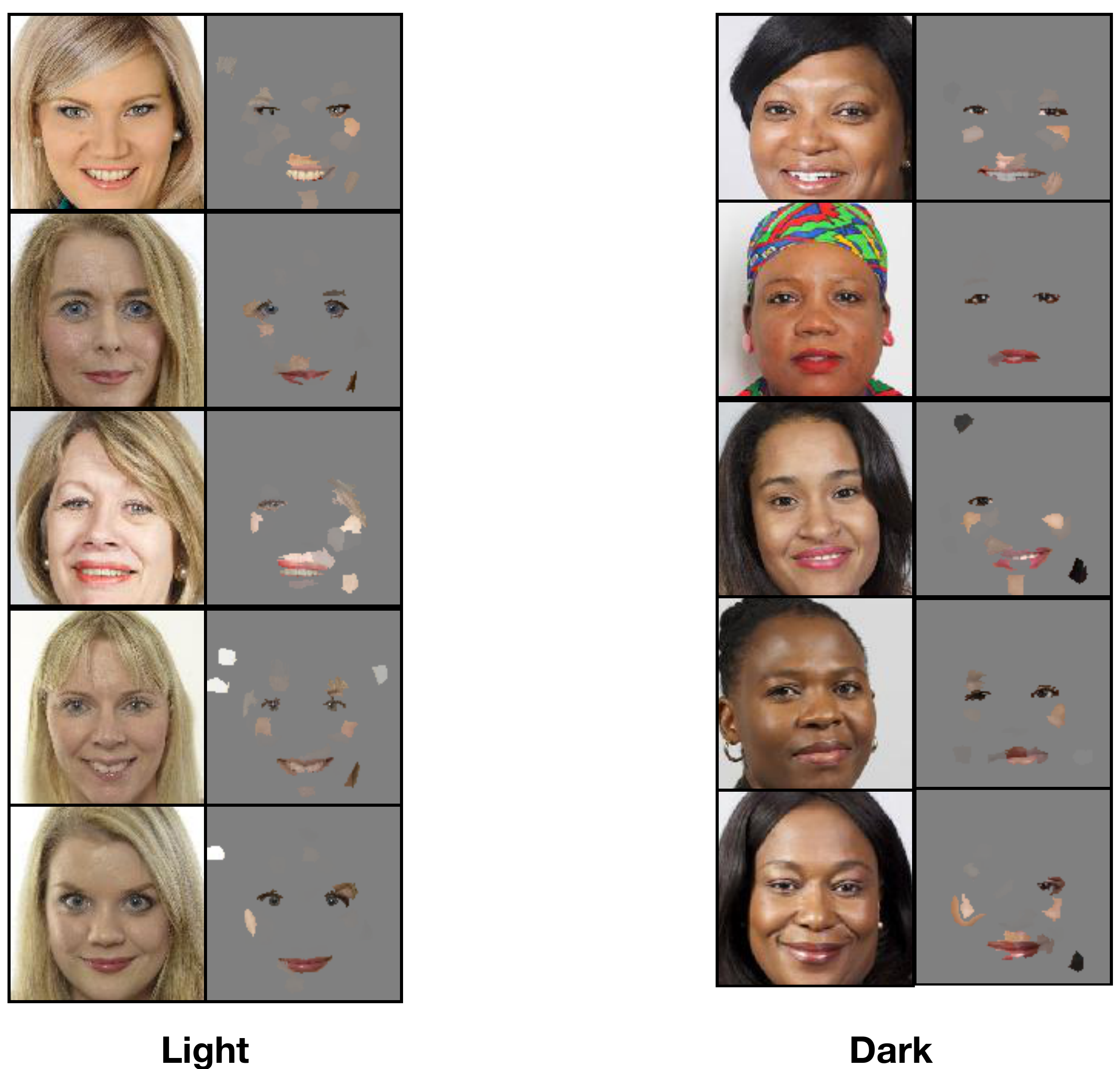}\\
\subcaption{Females.}\label{fig:contrastive_female_all}
\end{minipage}\\
\begin{minipage}[b]{0.4\textwidth}
\includegraphics[width=0.95\columnwidth]{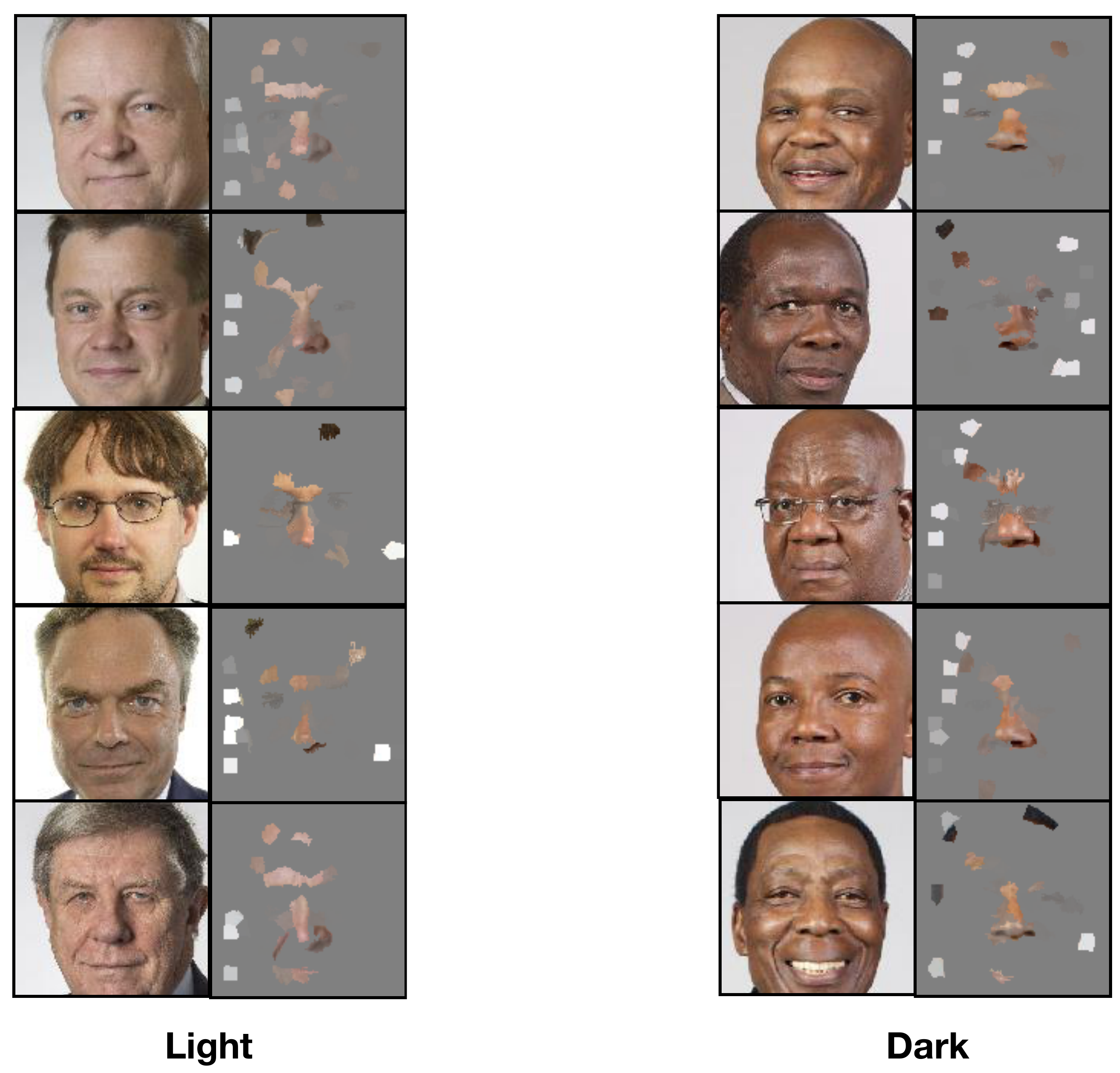}\\
\subcaption{Males.}\label{fig:contrastive_male_all}
\end{minipage}%
\caption{Sufficient explanations for sample females and males in PPB* dataset.}\label{fig:contrastive_all}
\end{figure}

\begin{figure}
\begin{minipage}[b]{0.25\textwidth}
\includegraphics[width=0.9\columnwidth]{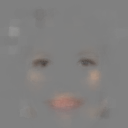}\\
\subcaption{Female.}\label{fig:contrastive_female_avg}
\end{minipage}%
\begin{minipage}[b]{0.25\textwidth}
\includegraphics[width=0.9\columnwidth]{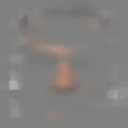}\\
\subcaption{Male.}\label{fig:contrastive_male_avg}
\end{minipage}%
\caption{Average sufficient explanations for all females and males in PPB* dataset.}\label{fig:contrastive_avg}
\end{figure}

What do gender classifiers look at, if not skin type and hair length?
One reason why this sort of question is so challenging to answer is the high dimensionality of facial images, but another reason is the complexity of ML models.
State-of-the-art gender classifiers typically use neural networks, whose interpretability is challenging and currently an active area of machine learning research~\cite{montavon2017methods}.
One class of interpretability methods explains a neural network's decision on each image~\cite{bach2015pixel,selvaraju2017grad,simonyan2013deep,lei2016rationalizing}.
The challenge is that superfluous features that do not actually contribute (or could even negatively contribute) to the classification decision can often be highlighted.

We seek \textit{minimal sufficient explanations} for a classification decision in an image.
In other words, what are the minimal features in an image that, by themselves, would be classified as female/male?
We use the recently proposed contrastive explanations method~\cite{dhurandhar2018explanations} to answer this question, particularly looking at finding pertinent positives.

\begin{procedure}
The \textbf{contrastive explanations method} takes as input an image $X$, which is classified in category $k$, and a (possibly neural network) model $f$ that maps images to logits on the classification decision.
It selects a ``pertinent positive explanation'' $\Delta(X)$ as a solution to the following optimization problem:
\begin{align}\label{eq:ppopt}
\min_{\Delta} c \cdot f_{\kappa}(X,\Delta) + \beta ||\Delta||_1 + ||\Delta||_2^2 ,
\end{align}
where $f_{\kappa}(X,\Delta) := \max\{\max_{k' \neq k} f(\Delta)_{k'} - f(\Delta)_k, -\kappa \}$, and the parameters $(c,\beta,\kappa)$ are regularization hyperparameters.
\end{procedure}

Effectively, the contrastive explanations method selects as simple an explanation as possible (where simplicity is measured by the elastic net regularizer~\cite{zou2005regularization}), subject to the classification decision of the original image being preserved.
This method has been validated on the MNIST hand-written digits dataset as well as an MRI dataset~\cite{dhurandhar2018explanations}, but has not yet been used to provide post hoc explanations in face classification tasks.

We set parameters $\kappa = 10$ and $\beta = 0.1$ and searched over the space of hyperparameters for $c$. 
 We applied the contrastive explanations method to all images in the PPB* dataset and used the customized neural network classifier described in Section~\ref{sec:models} as the function $f$.
Examples of contrastive explanations, effectively values of $\Delta(X)$ for \textit{correctly predicted} females and males in the PPB* dataset are presented in Figure~\ref{fig:contrastive_all}.
We can see in Figure~\ref{fig:contrastive_female_all} that the lips, eyes and cheeks show up very prominently as a sufficient explanation for a female classification --- in particular, female lips look pink/red in color, and cheekbones are more prominent.
This could be a result of celebrities, who wear more prominent makeup in photographs than the average human population, in training datasets.
In Figure~\ref{fig:contrastive_male_all}, we see that the nose and forehead area are highlighted as \textit{sufficient explanations} for a male classification.
These are clearly consistent patterns noticed across \textit{correctly classified} females and males.
The average sufficient explanation masks for females and males are presented in Figure~\ref{fig:contrastive_avg}.

Work in visual perception~\cite{brown1993gives,burton1993s} has shown that humans can adeptly classify the gender of a face using certain facial features in isolation.
It is interesting to see that ML models are also able to do this.
It is particularly interesting that a few of the dark females in Figure~\ref{fig:contrastive_female_all} have underrepresented skin type and hair pattern, and are yet classified correctly, probably largely due to their lip and cheek patterns.

However, the questions of racial \textit{and} gender bias are still relevant for facial features in isolation.
It has been statistically established that the facial structure of European- and African-descent females, including cheekbones and lips, is different~\cite{zhuang2010facial}.
So we cannot expect the performance of ML algorithms trained primarily on European-descent females to generalize to African-descent females, even if such simple facial features form a sufficient explanation.
And lip makeup alone constitutes a simplistic explanation of a female face, furthering a gender stereotype.

\section{Discussion and Future Work}

We rigorously tested the influence of various features on the gender classification task.
First, we showed that gender classification is relatively stable to variations in skin type and thus the skin type \textit{by itself} has a minimal effect on the classification decision.
Second, we observed unequal performance on females with varying hair length and tested the performance of a classifier that ignores hair information.
We saw that the unequal performance across gender and skin tone persists, suggesting that facial features other than skin type and hair pattern are behind the phenomenon.
Finally, using the contrastive explanations method, we identified red/pink lips, cheeks and eyes; and nose and forehead as sufficient facial features for a classification decision to be female or male respectively.

We began this research with the aim of developing invariant or equivariant face classifiers that would ignore skin type completely and thereby have equal accuracy across groups.  Such an approach would preclude the need for a high level of diversification in training datasets.  However, our mathematically-oriented analysis using the low-dimensional skin type group revealed that high-performing gender classifiers are already invariant to skin type.  Moreover, we showed that the classifiers are sensitive to a host of facial features that are not easily considered in isolation.  To solve the problem of unequal performance, we require diverse training datasets that represent humanity across many dimensions of identity, starting but not ending with ethnicity.

Many questions remain as to how exactly to go about diversifying training data as even ethnicity does not fully encapsulate an individual's identity. The contrastive explanations for female images consist of stereotypical attributes like lip makeup, and thus gender stereotypes commonly used by humans are confirmed in machine learning algorithms.
This is a parallel issue to the issue of bias in skin type and correlated attributes: while females and males are balanced in training data, they are stereotypical females and males from the \emph{celebrity} population.
Informally speaking, we would expect the appearance of the general population of females and males alike to be quite different.
We suggest that a good training dataset should diversify not only across ethnicity, but also across profession, cultural norms, and economic status, to capture a \emph{truly} global population.
Collecting such a dataset while controlling for image quality is a difficult, but necessary task.

As a parallel effort, it would also be interesting to examine the potential of decoupled classification on demographic groups, which along with task transfer learning has been shown to mitigate biases in classification of other facial attributes across race and gender~\cite{RyuAM2018}.

Finally, the perspectives presented here are limited to the problem of binary gender classification from visual data, itself a flawed problem especially when considering various non-binary gendered individuals.
The community needs to move beyond the binary gender construct in future work.

\section*{Acknowledgments}

This work was conducted under the auspices of the IBM Science for Social Good initiative. The authors thank Joy Buolamwini, Pin-Yu Chen, Amit Dhurandhar, Michele Merler, and Karthikeyan Natesan Ramamurthy for comments and assistance. 

\bibliographystyle{ieee}
\bibliography{references_cvpr}

\end{document}